\documentclass{article} % For LaTeX2e
\usepackage{iclr2025_conference,times}
\usepackage{graphicx}
\usepackage{algorithm}
\usepackage{algorithmic}
\usepackage{amssymb}
\usepackage{amsmath}
\usepackage{color, xcolor}
\usepackage{multirow}
\usepackage{booktabs}
\usepackage{subfigure}
\newtheorem{proof}{\indent Proof}[section]
\newtheorem{theorem}{\indent Theorem}[section]

\newtheorem{proposition}[theorem]{\indent Proposition}

\usepackage{amsmath,amsfonts,bm}

\def\eqref#1{equation~\ref{#1}}
\def\1{\bm{1}}

\DeclareMathAlphabet{\mathsfit}{\encodingdefault}{\sfdefault}{m}{sl}
\SetMathAlphabet{\mathsfit}{bold}{\encodingdefault}{\sfdefault}{bx}{n}

\usepackage{hyperref}
\usepackage{url}
\usepackage{wrapfig}

\usepackage{authblk}

\title{\centering Iterative Refinement of Flow Policies \\ in Probability Space for Online \\ Reinforcement Learning}

\author[1,2,3]{\textbf{Mingyang Sun}}
\author[1,2]{\textbf{Pengxiang Ding}}
\author[1,4]{\textbf{Weinan Zhang}}
\author[2*]{\textbf{Donglin Wang}}

\affil[]{Zhejiang University, 
\textsuperscript{2}Westlake University}

\affil[3]{Shanghai Innovation Institute, \textsuperscript{4}Shanghai Jiao Tong University}

\affil[ ]{\texttt{sunmingyang@westlake.edu.cn}}

\iclrfinalcopy 

\begin{document}

\maketitle

\begin{abstract}
While behavior cloning with flow/diffusion policies excels at learning complex skills from demonstrations, it remains vulnerable to distributional shift, and standard RL methods struggle to fine-tune these models due to their iterative inference process and the limitations of existing workarounds. In this work, we introduce the Stepwise Flow Policy (SWFP) framework, founded on the key insight that discretizing the flow-matching inference process via a fixed-step Euler scheme inherently aligns it with the variational Jordan–Kinderlehrer–Otto (JKO) principle from optimal transport. SWFP decomposes the global flow into a sequence of small, incremental transformations between proximate distributions. Each step corresponds to a JKO update, regularizing policy changes to stay near the previous iterate and ensuring stable online adaptation with entropic regularization. This decomposition yields an efficient algorithm that fine-tunes pre-trained flows via a cascade of small flow blocks, offering significant advantages: simpler/faster training of sub-models, reduced computational/memory costs, and provable stability grounded in Wasserstein trust regions. Comprehensive experiments demonstrate SWFP's enhanced stability, efficiency, and superior adaptation performance across diverse robotic control benchmarks.
\end{abstract}

\section{Introduction}

Robotic control has rapidly evolved from hand-crafted feedback systems to data-driven, end-to-end learning paradigms. Modern tasks demand not only precise execution of complex behaviors but also robust adaptability to environmental uncertainties. Imitation Learning (IL) has emerged as a dominant approach, with recent advances in Behavior Cloning (BC)—such as diffusion policies \cite{chi2023diffusion, DBLP:conf/corl/KeGF24, DBLP:conf/rss/ZeZZHWX24} and action chunking \citep{DBLP:conf/rss/ZhaoKLF23} robots to acquire sophisticated, long-horizon skills from demonstration data. Notably, generative policies based on diffusion or flow matching offer compelling advantages: they natively handle multi-modal action distributions, scale effectively in high-dimensional spaces, and achieve training stability through denoising and score-matching techniques.

% \textcolor{blue}{\% Speed, Optimization}
Despite these innovations, BC methods remain fundamentally vulnerable to distributional shift—where deviations from training states induce compounding errors \citep{ross2010efficient}. Reinforcement Learning (RL) offers a principled path for policy improvement through online interaction~\citep{chen2024aligning, wagenmaker2025steering,dppo}, while learning policies via RL from scratch is often too inefficient for practical use. Unfortunately, fine-tuning expressive generative policies presents unique challenges. Unlike Gaussian policies, diffusion or flow models require iterative inference (often tens of steps),  which fundamentally breaks the single-step gradient path required by standard policy-gradient or Q-learning algorithms. Existing workarounds—weighted regression\citep{lu2023contrastive}, implicit reparameterization\citep{ada2024diffusion}, or rejection sampling\citep{idql}—either introduce large gradient variance or lack a rigorous interpretation in terms of distribution optimization, leaving their stability and convergence properties fundamentally unclear.

In this work, we take steps towards addressing this challenge and seek to understand how we can utilize online interaction to adapt pre-trained flow policies in a simple and efficient way. 
Our core insight arises from reinterpreting the fixed-step Euler scheme of flow matching: the multi-step transport process from a noise distribution to an optimized action distribution naturally corresponds to traversing a Wasserstein gradient flow~\citep{santambrogio2017euclidean, zhang2018policy} in probability space. This perspective allows us to leverage powerful tools from optimal transport. Specifically, we rigorously align this process with the Jordan–Kinderlehrer–Otto (JKO) scheme~\citep{Jordan_Kinderlehrer_Otto_1998, mokrov2021large, xie2025flow}, a variational principle from optimal transport that provides theoretical guarantees for stable, incremental policy updates. By decomposing the global flow into a sequence of incremental transformations, each JKO step regularizes policy updates to stay near the previous iterate, mitigating instability while enabling efficient optimization.

We instantiate this concept in \textbf{Stepwise Flow Policy (SWFP)}, a stable algorithm that fine-tunes pretrained flow policies through a cascade of small, specialized flow blocks. Unlike monolithic flow matching, which directly bridges potentially distant noise and target distributions, SWFP structures the transport into localized steps where consecutive distributions are proximate. This decomposition yields three key advantages: (1) each sub-flow model in fixed step scheme is simpler to train and converges faster, (2) computational and memory costs are significantly reduced, and (3) the JKO alignment provides a theoretically grounded mechanism for stable policy improvement with entropic regularization.

\begin{itemize}
    \item We propose SWFP, the first framework that unifies Flow-Matching policies with the Jordan–Kinderlehrer–Otto (JKO) proximal operator, establishing a principled bridge between generative inference and regularized policy improvement in reinforcement learning.

    \item Building on the framework, we devise a practical algorithm that finetunes a pre-trained flow policy through a succession of short, easily trained flow blocks. The method retains the theoretical guarantees of Wasserstein trust regions while remaining practical in high-dimensional continuous-action settings.

    \item Comprehensive experiments demonstrating SWFP’s advantages, including enhanced stability, reduced computational overhead, and superior adaptation efficiency, across diverse robotic control benchmarks.
\end{itemize}

\section{Related Work}

\subsection{Generative Models for Decision Making} 

Generative models, particularly those based on diffusion \citep{ho2020denoising} and flow-matching \citep{lipman2022flow}, have become the state-of-the-art for learning complex robot behaviors. In imitation learning, they excel at capturing multi-modal action distributions from expert data, significantly outperforming standard Behavioral Cloning \citep{chi2023diffusion, braun2024riemannian}. They have also proven highly effective in offline reinforcement learning~\citep{he2023diffusion,wang2022diffusion,mao2024diffusion}, where models like Diffuser \citep{ajayconditional} use them for planning reward-conditioned trajectories. However, a crucial limitation persists across these applications: their iterative inference structure and large parameter counts make them fundamentally difficult to fine-tune efficiently using online RL~\citep{uehara2024understanding, clark2023directly, sun2025score}. Our work directly addresses this challenge of adapting powerful, pre-trained generative policies to new online information.

\subsection{RL with diffusion and flow models}

To bridge the gap between offline pre-training and online adaptation, several fine-tuning strategies have been explored. One line of work treats the multi-step denoising process as an MDP to apply policy gradients \citep{DBLP:conf/iclr/BlackJDKL24, fan2024reinforcement, dppo}, but this often leads to high variance and instability. A second, more indirect approach uses learned value functions to guide or filter the output of a frozen generative policy, for instance via advantage-weighted regression \citep{goo2022know}, rejection sampling \citep{idql}, or reformulating the objective as a supervised learning problem with return conditioning~\citep{chen2021decision,janner2022planning,ajayconditional}. While practical, these methods do not improve the underlying policy distribution itself. In contrast to these prior works that struggle with stability or are indirect, our method, SWFP, introduces a principled framework for performing direct, stable, and incremental updates to the flow policy's distribution. By grounding our algorithm in the Jordan–Kinderlehrer–Otto (JKO) scheme, we provide a theoretically sound mechanism for online flow policy improvement.

% As demonstration data are often limited, there have been many approaches proposed to improve the performance of diffusion-based policies. 
% One straightforward approach \citep{DBLP:conf/iclr/BlackJDKL24, fan2024reinforcement} involves framing diffusion denoising as a Markov Decision Process (MDP), which facilitates preference-aligned generation with policy gradient reinforcement learning. However, this approach often suffers from instability, limiting its practical applicability. \citep{dppo} introduced policy gradient loss on a two-layer MDP for direct diffusion policy fine-tuning, which mitigates this instability, but the method is architecture-specific and does not introduce closed-loop control. 
% Alternative approaches to integrating diffusion architectures with reinforcement learning (RL) include leveraging Q-function-based importance sampling~\citep{idql}, employing advantage-weighted regression~\citep{goo2022know}, or reformulating the objective as a supervised learning problem with return conditioning~\citep{chen2021decision,janner2022planning,ajayconditional}.
% Additionally, researchers have explored enhancing the denoising training objective by incorporating Q-function maximization~\citep{wang2022diffusion} and iteratively refining the dataset using Q-functions~\citep{yang2023policy}. 
% Another promising direction involves augmenting a frozen, chunked diffusion policy model with a residual policy trained through online RL, enabling improved performance without modifying the pre-trained diffusion model~\citep{ankile2024imitation}.

\section{Preliminaries}

\textbf{Flow matching.} 
% Flow matching is a method for generative modeling, specifically for training residual models \citep{Lipman2022FlowMF}. 
% Unlike standard diffusion models, which model dynamics via stochastic differential equations (SDEs), flow matching uses ordinary differential equations (ODEs) to describe a deterministic transformation (no additive noise) of a random sample from an initial distribution, thereby simplifying training and linking naturally to continuous normalizing flows: Flow matching allows the training of continuous normalizing flows in a simulation-free manner.
Continuous Normalizing Flows (CNFs) \citep{DBLP:conf/nips/ChenRBD18} considers the dynamic of the probability density function by probability density path \( p : [0, 1] \times \mathbb{R}^d \rightarrow \mathbb{R}_{\geq 0} \) which transmits between the data distribution \( p_1 \) and the initial distribution (e.g., Gaussian distribution) \( p_0 \). The flow \( \boldsymbol{\phi} : [0, 1] \times \mathbb{R}^d \rightarrow \mathbb{R}^d \) is constructed by a vector field \( \boldsymbol{v} : [0, 1] \times \mathbb{R}^d \rightarrow \mathbb{R}^d \) describing the velocity of the particle at position \( \boldsymbol{x} \), i.e., the ODE
\begin{equation}
\frac{\mathrm{d}}{\mathrm{d}t}\boldsymbol{\phi}_t(\boldsymbol{x}) = \boldsymbol{v}_t(\boldsymbol{\phi}_t(\boldsymbol{x})),
\end{equation}
where $\boldsymbol{\phi}_0(\boldsymbol{x}) = \boldsymbol{x}$.
In order to ensure that the vector field \( \boldsymbol{v} \) generates the probability density path \( p_t \), the following \textit{continuity equation}~(CE) \citep{Villani_2013} is required:
\begin{equation}
\frac{\mathrm{d}}{\mathrm{d}t} p_t(\boldsymbol{x}) + \text{div}\cdot \left[ p_t(\boldsymbol{x})\boldsymbol{v}_t(\boldsymbol{x}) \right] = 0, \quad \forall \boldsymbol{x} \in \mathbb{R}^d
\end{equation}

Given such a process, flow matching models learn a neural network $\boldsymbol{v}^\theta_t$ to learn the ground truth vector field $\boldsymbol{u}_t$ by minimizing their differences, i.e., $\mathcal{L}_{FM}(\theta) = \mathbb{E}_{t, p_t(\boldsymbol{x})}\|\boldsymbol{v}^\theta_t(\boldsymbol{x})-\boldsymbol{u}_t(\boldsymbol{x})\|^2_2$ with respect to the network parameter $\theta$. 

However, the original FM objective is generally intractable because the time-varying distribution $p_t(\boldsymbol{x})$ and the true vector field $\boldsymbol{u}_t(\boldsymbol{x})$ are often intractable. To address this, Conditional Flow Matching (CFM) has been proposed (Lipman et al., 2023), which simplifies the task by conditioning the flow on target samples to derive a tractable objective. The CFM loss is defined as follows:
\begin{equation}
\mathcal{L}_{\text{CFM}}(\theta) = \mathbb{E}_{\substack{t \sim \mathcal{U}(0,1), \\ \boldsymbol{x}_1 \sim q(\boldsymbol{x}_1), \\ \boldsymbol{x} \sim p_t(\boldsymbol{x}|\boldsymbol{x}_1)}} \left[ \left\| \boldsymbol{v}_\theta(t, \boldsymbol{x}) - \boldsymbol{u}_t(\boldsymbol{x} | \boldsymbol{x}_1) \right\| \right]^2 \tag{1}
\end{equation}
wherein $\boldsymbol{u}_t(\boldsymbol{x} | \boldsymbol{x}_1)$ becomes tractable by defining explicit conditional probability paths from $\boldsymbol{x}_0$ to $\boldsymbol{x}_1$, such as OT-paths or linear interpolation paths.

\noindent\textbf{Reinforcement Learning}
In this paper, we focus on policy learning in continuous action spaces. We consider a Markov Decision Process (MDP) defined by the tuple $(\mathcal{S}, \mathcal{A}, \mathcal{P}, r, \rho_0, \gamma)$, where $\mathcal{S}$ represents the state space, $\mathcal{A}$ is the continuous action space, $\mathcal{P} : \mathcal{S} \times \mathcal{S} \times \mathcal{A} \rightarrow [0, +\infty]$ is the probability density function of the next state $\boldsymbol{s}' \in \mathcal{S}$ given the current state $\boldsymbol{s} \in \mathcal{S}$ and the action $\boldsymbol{a} \in \mathcal{A}$, $r : \mathcal{S} \times \mathcal{A} \rightarrow [r_{\text{min}}, r_{\text{max}}]$ is the bounded reward function.

The standard RL aims to learn a policy that maximizes the expected cumulative reward: $\mathbb{E}_\pi[r(\boldsymbol{a})] = \int_\mathcal{A} r(\boldsymbol{a})\mathrm{d}\pi(\boldsymbol{a})$. 
Shannon entropy is often added as a regularization term to improve exploration and avoid early convergence to suboptimal policies. This gives us the entropy-regularised reward, which is a free energy functional, named by analogy with a similar quantity in statistical mechanics:
\begin{equation}\label{eq:rlobj}
J(\pi) = \int_\mathcal{A} r(\boldsymbol{a})\mathrm{d}\pi(\boldsymbol{a}) - \beta \int_\mathcal{A} \log\pi(\boldsymbol{a})\mathrm{d}\pi(\boldsymbol{a})
\end{equation}
Equation (\ref{eq:rlobj}) is often interpreted as a free energy functional by analogy with statistical mechanics, and it underlies several state-of-the-art algorithms such as soft Q-learning, soft Actor-Critic and entropy-regularized policy gradient methods~\citep{haarnoja2017reinforcement, nachum2017bridging, haarnoja2018soft}.

We are interested in the process of policy iteration, that is, finding a sequence of policies $(\pi_n)$ converging towards the optimal policy $\pi^*$. Policy iteration is often implemented using gradient ascent according to
\begin{equation}
\pi_{n+1} = \pi_n + \tau \nabla J(\pi_n)
\end{equation}
% Rearranging gives the implicit Euler method
% \begin{equation}
% \frac{\pi_{k+1} - \pi_k}{\tau} - \nabla J(\pi_{k+1}) = 0
% \end{equation}

In implicit Euler method, if integrated and interpreted as an $L^2$ regularized iterative problem, it is strictly equivalent to finding a solution to the proximal problem:
\begin{equation}
    \pi_{n+1} = \arg\min_{\pi}\frac{\|\pi-\pi_n\|^2}{2\tau} - J(\pi)
\end{equation}
Rather than just the $L^2$ distance between policies for constraining and regularization, one can envision the more general case of any policy distance $d$:
\begin{equation}\label{eq:d_obj}
    \pi_{n+1} = \arg\min_{\pi}\frac{d^2(\pi, \pi_n)}{2\tau} - J(\pi)
\end{equation}

% where $\mathcal{H}(\pi(\cdot|s_t)) = \mathbb{E}_{a_t \sim \pi(\cdot|s_t)} \left[ -\log \pi(a_t|s_t) \right]$, and $\beta$ is the temperature parameter that controls the trade-off between the entropy and reward terms. A higher value of $\beta$ drives the optimal policy to be more stochastic, which is advantageous for RL tasks requiring extensive exploration. In contrast, the standard RL objective can be seen as the limiting case where $\beta \rightarrow 0$.

\section{Method}
In this section, we first identify the difficulties in training of flow-based policy in the context of online RL. To mitigate this, we propose our core contribution, SWFP, an iterative flow using JKO scheme. Finally, we give a practical algorithm with soft critic and parallel block training.

\subsection{Flow as RL policy}
We begin by formalizing our flow-based policy framework. The core architecture employs flow matching in action space $\mathcal{A}$, parameterized by a state- and time-dependent vector field $\boldsymbol{v}_\theta(t, \boldsymbol{s}, \boldsymbol{a})$. The fundamental behavioral cloning objective is given by:
\begin{equation}\label{eq:flow_BC}
\mathcal{L}_{\text{FM}}(\theta) = \mathbb{E}_{\substack{\boldsymbol{s},\boldsymbol{a}_1 \sim \mathcal{D}, \\ \boldsymbol{a}_0 \sim \mathcal{N}(0, I_d), \\ t \sim \mathcal{U}(0,1)}} \left[ \left\| \boldsymbol{v}_\theta(t, \boldsymbol{s}, \boldsymbol{a}_t) - \boldsymbol{u}_t(\boldsymbol{a}_t|\boldsymbol{a}_1) \right\|_2^2 \right] 
\end{equation}
The state-dependent vector field generates a state-dependent flow $\boldsymbol{\phi}_\theta(t, \boldsymbol{s}, \boldsymbol{a}) : [0, 1] \times \mathcal{S} \times \mathbb{R}^d \rightarrow \mathbb{R}^d$. For $\boldsymbol{s} \in \mathcal{S}$ and $\boldsymbol{z} \in \mathbb{R}^d$, $\boldsymbol{\phi}_\theta(1, \boldsymbol{s}, \boldsymbol{z})$ maps the noise $\boldsymbol{z} = \boldsymbol{x}_0$ (sampled from the standard normal distribution $\mathcal{N}(0, I_d)$) to the action $\boldsymbol{a} = \mu_\theta(\boldsymbol{s}, \boldsymbol{z})$ by the ODE. $\mu_\theta(\boldsymbol{s}, \boldsymbol{z})$ is a deterministic function from $\mathcal{S} \times \mathbb{R}^d$ to $\mathcal{A}$, but serves as a stochastic policy $\pi_{\theta}(\boldsymbol{a}|\boldsymbol{s})$ from $\mathcal{S}$ to $\mathcal{A}$ due to the stochasticity of $\boldsymbol{z}$.

During online RL, optimizing the objective in (\ref{eq:rlobj}) with a maximum entropy constraint provides us with a framework for training stochastic policies. Speicially, we consider a general energy-based policies of the form $\pi(\boldsymbol{a}|\boldsymbol{s}) \propto \exp(-\epsilon(\boldsymbol{s}, \boldsymbol{a})/\alpha)$ that is able to model more complex distributions. Following the soft policy iteration algorithm, the policy is updated to fit the target max-entropy policy
\begin{equation}\label{eq:exp_maxent}
    \pi_{\mathrm{MaxEnt}}(\boldsymbol{a}|\boldsymbol{s}) \propto {\exp(Q^{\pi_{\mathrm{old}}}(\boldsymbol{s}, \boldsymbol{a}))}
\end{equation}
where $Q^{\pi_{\mathrm{old}}}(\boldsymbol{s}, \boldsymbol{a})$ is the converged result of soft Bellman update operator.

However, training flow-based policies in online reinforcement learning remains highly challenging due to two fundamental limitation. Firstly, the flow-matching objective (\ref{eq:flow_BC}) becomes intractable in online RL settings, as it requires samples from the entropy-regularized optimal policy (\ref{eq:exp_maxent})—which are unavailable during iterative policy improvement.
Secondly, while one might consider treating the reverse generative process as a policy parameterization and backpropagating gradients through the full sampling trajectory, this approach incurs prohibitive computational and memory overhead from recursive gradient computations.

These limitations motivate our novel training framework presented in subsequent sections, which maintains the expressiveness of flow-based policies while addressing these practical constraints.

\subsection{Iterative flow using JKO scheme}

\begin{figure}[t]
\centering
\includegraphics[trim = 50mm 10mm 40mm 90mm, clip, width=0.7\columnwidth]{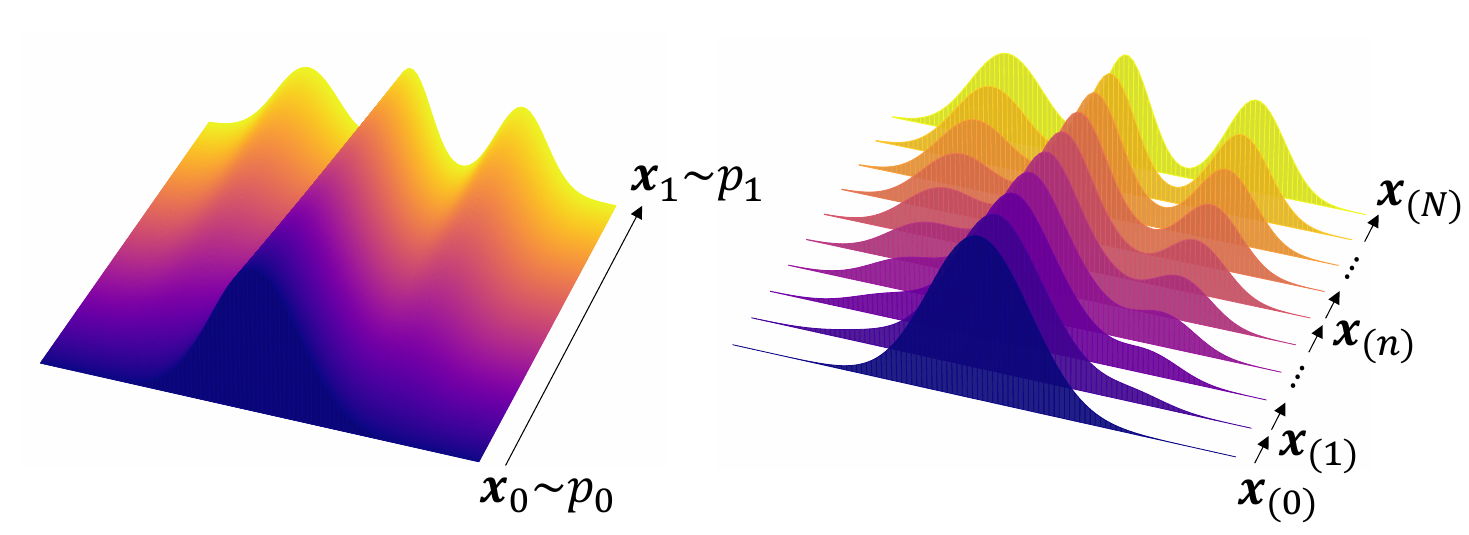} 
\caption{The illustration of continuous-time flow (left) and discrete-time block flow (right).  }
\label{fig:blockflow}
\end{figure}

In the continuous-time flow trained, let $\boldsymbol{x}_t$ be the ODE solution trajectory satisfying $\dot{\boldsymbol{x}}_t = \boldsymbol{v}_\theta(t, \boldsymbol{x}_t)$ on $[0, T]$, the flow mapping can written as:
\begin{equation}\label{eq:ctf}
    F_\theta(\boldsymbol{x}) = \boldsymbol{x} + \int^T_0 \boldsymbol{v}_\theta(\boldsymbol{x}_t, t) \mathrm{d}t,\quad \boldsymbol{x}_1 = \boldsymbol{x}.
\end{equation}
which induces an instantaneous change-of-variable formula of parametric log-density: $\log p_t(\boldsymbol{x}_t) = \log p_0(\boldsymbol{x}_0) - \int^t_0\nabla\cdot \boldsymbol{v}_\theta(\boldsymbol{x}_t, s)\mathrm{d}s$.
Although the formulation is continuous, any practical evaluation of~(\ref{eq:ctf}) inevitably relies on a \emph{discrete} time grid; indeed, a single residual block can be viewed as a forward–Euler step that integrates the same ODE for one unit time.
% This observation motivates us to \emph{embed} continuous–time expressiveness inside a discrete-time architecture.
Conversely, this observation motivates us to utilize the benefit of continuous-time NF (neural ODE) inside the discrete-time NF framework by setting the $n$-th block $f_n$ to be a neural ODE on a sub-interval of time.

Specifically, let the time horizon $[0,T]$ be discretized into $N$ subintervals $[t_{n-1}, t_n]$ and $\boldsymbol{x}_t$ solves the ODE with respect to the velocity field $v_\theta(\boldsymbol{x}_t, t)$. The $n$-th block mapping (associated with the subinterval $[t_{n-1}, t_n]$) is defined by the integral flow:
\begin{equation}\label{eq:blockFlow}
\boldsymbol{x}_{(n)} = \boldsymbol{x}_{(n-1)} + \int_{t_{n-1}}^{t_n} \boldsymbol{v}_\theta(\boldsymbol{x}_t, t) \mathrm{d}t, \quad \boldsymbol{x}_{t_{n-1}} = \boldsymbol{x}_{(n-1)}.
\end{equation}
Equation~\ref{eq:blockFlow} inherits the analytically tractable change-of-variables formula of CNFs—one integrates $\nabla \boldsymbol{v}_\theta$  over the same interval.

Obviously, standard FM training treats the entire trajectory on $[0,T]$ as a monolithic map and optimizes all $N$ blocks jointly with a single objective (typically maximum likelihood), see Fig.~\ref{fig:blockflow}(left). While by adopting a flow sub-network inside each residual block, one can design a discrete-time flow model that is free-form, automatically invertible (by using small time step to ensure sufficiently accurate numerical integration of the ODE such that the ODE trajectories are distinct), and enjoys the same computational and expressive advantage as continuous-time flow policy.
If using $\boldsymbol{v}_\theta$ on $[t_{n-1}, t_{n}]$, the $n$-th block can potentially be trained independently and progressively, meaning that only one block is trained at a time and the $n$-th block is trained only after the previous $(n-1)$ blocks are fully trained and fixed. We call such flow implementing the iterative steps the iterative flow, and the key of our SWFP is to design a step-wise loss to train each block.

To solve the above problem, we first model the continuous-time diffusion process by a partial differential equations (PDE), i.e., the Fokker-Planck equation,
\begin{equation}\label{eq:FP}
    \frac{\partial \rho(t, \boldsymbol{x})}{\partial t} = \nabla \cdot (\nabla U(\boldsymbol{x}) \rho(t, \boldsymbol{x})) + \beta \nabla^2 \rho(t, \boldsymbol{x}),
\end{equation}
describes the time evolution of the distribution $\rho$ of a set of particles undergoing drift and diffusion. 
Motivated by the work of Jordan, Kinderlehrer, and Otto~\citep{Jordan_Kinderlehrer_Otto_1998, mokrov2021large}, related diffusion processes to energy-minimizing trajectories in the Wasserstein space, providing a discrete-time counterpart of the transport process, the JKO scheme.

The classical JKO scheme computes a sequence of distributions $\rho_n, n = 0, 1, \dots$ by 
\begin{equation}\label{eq:JKO}
    \rho_{n+1} = \arg\min_{\rho\in\mathcal{P}}\mathcal{E}(\rho) + \frac{1}{2\tau}\mathcal{W}^2_2(\rho_n, \rho),
\end{equation}
starting from $\rho_0\in\mathcal{P}_2$, where $\tau > 0$ controls the step size. $\mathcal{E}$ is an energy functional. The Fokker-Plank equation~(\ref{eq:FP}) results from the continuous-time limit (i.e., $\tau\rightarrow 0$) of the JKO scheme for the free energy and describes external potentials and local self-interactions:
\begin{equation}\label{eq:fe}
    \mathcal{E}(\rho) := \int \rho \log \rho \, dx + \int U(x) \, d\rho(x)
\end{equation} 
which is also expressed as the Kullback-Leibler (KL) divergence to the target distribution $q \propto e^{-U(x)}$.

Strictly speaking, the minimization in eq.~(\ref{eq:JKO}) is over the Wasserstein-2 space of the density $\rho$, which, in the $n$-th JKO flow block will apply to the pushforwarded density by the mapping in the $n$-th block. Specifically, let $F_{n,\theta}$ denote the forward mapping in the $n$-the block, i.e., $\boldsymbol{x}_n = F_{n, \theta}(\boldsymbol{x}_{n-1}, n-1)$ parameterized by $\boldsymbol{v}_\theta$ over $t \in [t_{n-1}, t_n]$. Denoting the marginal distribution of $\boldsymbol{x}{(n)}$ by $p_n$, with $p_N = p_1$ (where $\boldsymbol{x}_1$ follows the data distribution), we obtain the density transformation:
\begin{equation}
    p_n = (F_{n, \theta})_{\#}p_{n-1}.
\end{equation}

\noindent\textbf{Policy Improvement via JKO Iterations} For maximum entropy reinforcement learning, we instantiate the JKO scheme by setting the target distribution (\ref{eq:JKO}) to the energy-based policy from  (\ref{eq:exp_maxent}). Substituting $U(\boldsymbol{a}) \equiv -Q(\boldsymbol{s},\boldsymbol{a})/\alpha$ into the energy functional yields the policy optimization objective of the iterative flow policy:
\begin{align}\label{eq:jkopi}
     \pi_{n+1} &= \arg\min_{\pi}\frac{1}{2\tau}\mathcal{W}^2_2(\pi_n, \pi) 
     + \int\pi\log\pi\mathrm{d}\boldsymbol{a} + \int(-Q(\boldsymbol{s},\boldsymbol{a})/\alpha)\pi\mathrm{d}\boldsymbol{a}
\end{align}

This objective is minimized over the parameters $\theta$ of the flow map $F_\theta$, which transports the distribution $\pi_n$ to $\pi_{n+1}$. The loss has a clear interpretation: it seeks a new policy $\pi_{n+1}$ that minimizes the Max-Entropy objective while staying proximal to the previous policy $\pi_n$ in the Wasserstein space. The Wasserstein term serves as the policy distance $d$ in (\ref{eq:d_obj}) to regularize the ``amount of movement" from the current density $p_{\pi_{n-1}}$ by the transport map $F_{n,\theta}$. We provide more analysis in Appendix~\ref{app:ana}.
% The $\mathcal{W}_2^2$ term thus acts as a \textit{Wasserstein Trust Region} regularization.

\begin{proposition}\label{p1}
    For an iterative  flow with the energy functional defined in (\ref{eq:jkopi}), $\pi(\boldsymbol{a}|\boldsymbol{s})$ converges to $p_{s,\pi}(\boldsymbol{a}) \propto e^{Q(\boldsymbol{a},\boldsymbol{s})}$ in the infinite-time limit with $Q(\boldsymbol{a},\boldsymbol{s})$ satisfying the following modified Bellman equation:
\[
Q(\boldsymbol{a}, \boldsymbol{s}) = r(\boldsymbol{a}, \boldsymbol{s}) + \gamma \mathbb{E}_{\boldsymbol{s}' \sim \rho_\pi} \left[ V_\pi(\boldsymbol{s}') - \mathcal{H}(\pi(\cdot|\boldsymbol{s}')) \right]
\]
where $V_\pi(\boldsymbol{s}') \triangleq \log \int_{\mathcal{A}} \exp(Q(\boldsymbol{a}, \boldsymbol{s}')) \mathrm{d}\boldsymbol{a}$.
\end{proposition}
The proof is given in Appendix~\ref{p1}. 
% The per-step training objective (\ref{eq:jkopi}) can be viewed as the addition of the max entropy objective $J(\pi)$ (closeness of the pushforwareded density to target $p_{\pi}(\boldsymbol{a})$) and the Wasserstein term (the squared $\mathcal{W}_2$ distance between the pushforwareded density and the current density). 
This result provides theoretical grounding that by iteratively minimizing the JKO objective, SWFP converges to the optimal MaxEnt policy, ensuring stability through the intrinsic Wasserstein trust region imposed by the JKO operator. This step-wise form essentially decomposes the pushforward process of a flow model regarded as a policy into finite blocks of policy probability transport $(F_{n,\theta})_\#(\pi_{n-1})$.

\subsection{Stepwise policy optimization for flow matching}

% We have proposed training a sequence of flow-matching sub-models over $[0, T]$, each optimized to transport between consecutive distributions in the JKO scheme.  Our approach decomposes the global transport into localized steps where each sub-flow matches the terminal density evolved by the JKO scheme at its time step. policies form a Riemannian manifold on the space of probability measures. The manifold structure is determined by the expected total reward, and the geodesic length between two elements (policy distributions) is defined as the standard second-order Wasserstein distance. With convex energy functionals (defined below), searching for an optimal policy reduces to running SGD on the manifold of probability measures. 

From the JKO perspective, a policy $\pi$ is viewed as a point on the Wasserstein manifold of probability measures, where optimization involves following the steepest ascent of reward while paying a transportation cost measured by the W2 metric.  Rather than learning one monolithic flow that jumps directly from the behaviour distribution to the optimum, we decompose the trajectory into a sequence of short, local moves. Each flow-matching sub-model is trained to transport the current policy distribution to the next JKO iterate, so the global optimisation is realised as a chain of conservative, well-conditioned steps. This stepwise decomposition maintains proximal distributions between consecutive policy iterates,  enabling stable and efficient optimization. Building upon this geometric foundation, we now present a practical RL algorithm (see Algorithm~\ref{alg:traingpolicybrief} in Appendix).

\noindent\textbf{Soft critic for target policy}
Equation~(\ref{eq:jkopi}) adopts the JKO scheme to optimize the policy $\pi$ by particle approximation, i.e., $\pi \propto \frac{1}{M}\sum^M_{i=1}\delta_{\boldsymbol{a}^{i}}$. 
However, it is difficult due to the infinite time horizon and the unknown reward function $r(\boldsymbol{s},\boldsymbol{a})$ when calculating $Q(\boldsymbol{a}^{i}, \boldsymbol{s})$. 
To address this, we approximate the soft Q-function, $Q(\cdot, \boldsymbol{s})$, with a deep neural network $Q_{\boldsymbol{s}}^{\varphi}(\boldsymbol{s}, \boldsymbol{a})$ parametrized by $\varphi$, i.e., $p_{\boldsymbol{s}, \pi}(\boldsymbol{a}) \propto e^{Q_{\boldsymbol{s}}^{\varphi}(\boldsymbol{s}, \boldsymbol{a})}$. 
The neural network $Q_{\boldsymbol{s}}^{\varphi}(\boldsymbol{s}, \boldsymbol{a})$ naturally leads to a soft approximation of the standard Q-function according to Proposition~\ref{p1}.

We optimize the Q-network using the Bellman error as in the soft-Q learning setting. Specifically, in each iteration, we optimize the following objective function:
\begin{equation}
J_Q(\varphi) \triangleq \mathbb{E}_{\boldsymbol{s}_t \sim q_{\boldsymbol{s}_t}, \boldsymbol{a}_t \sim q_{\boldsymbol{a}_t}} \left[ \frac{1}{2} \left( \hat{Q}_s^{\bar{\varphi}}(\boldsymbol{s}_t, \boldsymbol{a}_t) - Q_s^\varphi(\boldsymbol{s}_t, \boldsymbol{a}_t) \right)^2 \right],
\end{equation}
where \( q_{\boldsymbol{s}_t} \) and \( q_{\boldsymbol{a}_t} \) are arbitrary distributions with support on \( \mathcal{S} \) and \( \mathcal{A} \), respectively; \begin{equation}
    \hat{Q}_s^{\bar{\varphi}}(\boldsymbol{s}_t, \boldsymbol{a}_t) = r(\boldsymbol{s}_t, \boldsymbol{a}_t) + \gamma \mathbb{E}_{\boldsymbol{s}_{t+1} \sim \rho_\pi} \left[ V_s^{\bar{\varphi}}(\boldsymbol{s}_{t+1}) \right]
\end{equation} is the target Q-value, with 
\begin{equation}
    V_s^{\bar{\varphi}}(\boldsymbol{s}_{t+1}) = \log \mathbb{E}_{q_{\boldsymbol{a}'}} \left[ \frac{\exp(Q_s^{\bar{\varphi}}(\boldsymbol{s}_{t+1}, \boldsymbol{a}'))}{q_{\boldsymbol{a}'}(\boldsymbol{a}')} \right] - \mathcal{H}(q_{\boldsymbol{a}'});
\end{equation} \( \bar{\varphi} \) represents the parameters of the target Q-network, as used in standard deep Q-learning. \( q_{\boldsymbol{a}_t} \) can be set to the distribution induced by the sampling network, built upon the flow model framework. Then, given $Q^{\varphi}_s$, we could adopt the particle approximation with the the JKO scheme to optimize the policy.

\begin{wrapfigure}{r}{0.5\textwidth}
\centering
\includegraphics[trim = 50mm 60mm 130mm 85mm, clip, width=0.5\columnwidth]{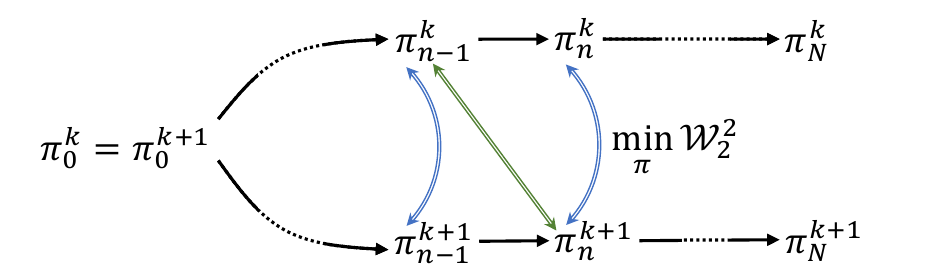} 
\caption{Illustration of SWFP's parallel block training. Each horizontal chain of black arrows represents the forward composition of flow blocks within a training epoch. Double-headed arrows indicate the Wasserstein distance minimization objectives: $\mathcal{W}^2_2(\pi^{k+1}_{n-1}, \pi^{k}_{n-1})$ and $\mathcal{W}^2_2(\pi^{k}_{n-1}, \pi^{k+1}_{n})$ during optimization of block $n$.
}
\label{fig:OTM}
\end{wrapfigure}

% \begin{figure}[h]
% \begin{center}
% %\framebox[4.0in]{$\;$}
% \fbox{\rule[-.5cm]{0cm}{4cm} \rule[-.5cm]{4cm}{0cm}}
% \end{center}
% \caption{Sample figure caption.}
% \end{figure}

\noindent\textbf{Parallel block training} 
% Although the final objective is $J(\pi)$, 
Under the JKO scheme, we specify a discrete time step schedule $\gamma_n = 1 / N$ for simplicity on the time interval $[0, 1]$. Suppose the data distribution $p(a_1)$ has a regular density $q$.
Starting from the noise distribution $p_0$ (when $n = 1$), we need recursively construct target density $p_{n} = F_{n, \theta, \#}p_{n-1}$, because of the dependency of $\pi_n$ on $\pi_{n-1}$ in (\ref{eq:jkopi}). In the first iteration, we train the first velocity $\boldsymbol{v}_\theta(\boldsymbol{a}, 0)$. After the first block is trained, the particle positions are updated using the learned transport map $F_{1, \theta}$.  
In the next iteration, we train the velocity field $v_\theta(\boldsymbol{a}, t_1), t_1 = \gamma*1$, and the initial position of the particles are $\boldsymbol{a}^{(i)}(t_1)$ which have been computed from the previous iteration. This procedure continues for $n = 1, 2, \dots , N$ for $N$ steps. This sequential training approach would clearly be inefficient and would require either independent or specially designed networks to mitigate catastrophic forgetting. 
Therefore, we propose parallel block training by introducing the triangle inequality of the Wasserstein distance: $\mathcal{W}^2_2(\pi^{k+1}_{n-1} , \pi^{k+1}_n) \leq  \mathcal{W}^2_2(\pi^{k+1}_{n-1}, \pi^{k}_{n-1}) + \mathcal{W}^2_2(\pi^{k}_{n-1}, \pi^{k+1}_{n})$. Specifically, during training at the $k+1$-th epoch, we utilize the old particle positions to compute the iterative process while additionally introducing the regularization term $W^2_2(\pi^k_{n}, \pi^{k+1}_{n})$:
\begin{align}\label{eq:jkopi_2}
     \pi^{k+1}_{n+1} &= \arg\min_{\hat\pi}\frac{1}{2\tau}\left(\mathcal{W}^2_2(\pi^{k}_n, \hat\pi) + \mathcal{W}^2_2(\pi^k_{n+1}, \hat\pi)\right) \notag \\
     &+ \int\hat\pi\log\hat\pi\mathrm{d}\boldsymbol{a} + \int(-Q(\boldsymbol{s},\boldsymbol{a})/\alpha)\hat\pi\mathrm{d}\boldsymbol{a} \triangleq J_{\pi_n}
\end{align}

The overall transport from the initial policy \(\pi_{0}\) to the target policy is decomposed into \(N\) consecutive FM blocks \(\{F_{n,\theta}\}_{n=1}^{N}\). At block $n$ the particles $\{\boldsymbol{a}^{i}_{(n)}\}^M_{i=1}$ is pushed forward by $\boldsymbol{a}^{i}_{(n+1)} = F_{n,\theta}(\boldsymbol{a}^{i}_{(n)}, \boldsymbol{s}, n)$.

Using the chain rule,
\[
\frac{\partial J_{\pi_n}^\theta}{\partial \theta}
=\;
\mathbb{E}\!\left[
  \frac{\partial J_{\pi_n}^\theta}{\partial \boldsymbol{a}^{i}_{(n)}}
  \,\frac{\partial \boldsymbol{a}^{i}_{(n)}}{\partial \theta}
\right],
\]
so \(\theta\) can be updated with standard stochastic gradient descent; \(\partial \boldsymbol{a}^{i}_{(n)}/\partial\theta\) is obtained by back-propagating through \(F_{n,\theta}\), and \(\partial J_{\pi}^\theta/\partial \boldsymbol{a}^{i}_{(n)}\) follows directly from the loss.  
Repeating this block-wise procedure yields controlled step sizes in Wasserstein space and ensures monotonic improvement of the entropy-regularised return.

\section{Experimental Results}

To comprehensively assess the proposed SWFP, we organize the experiments along four complementary axes that move from toy and RL benchmark comparison to a systematic hyper-parameter study. We begin with low-dimensional toy data to make convergence behavior tangible; next, we investigate how quickly the model can adapt in an online, interactive setting; third, we test whether a network pre-trained on large offline datasets can be retargeted to new objectives with minimal effort; and, finally, we analyze the sensitivity of performance to the number of JKO blocks.

\subsection{Foundational Toy Experiments}

\begin{wrapfigure}{r}{0.5\textwidth}
\centering
\subfigure{\includegraphics[trim = 0mm 0mm 0mm 0mm,  clip,width=0.48\linewidth ]{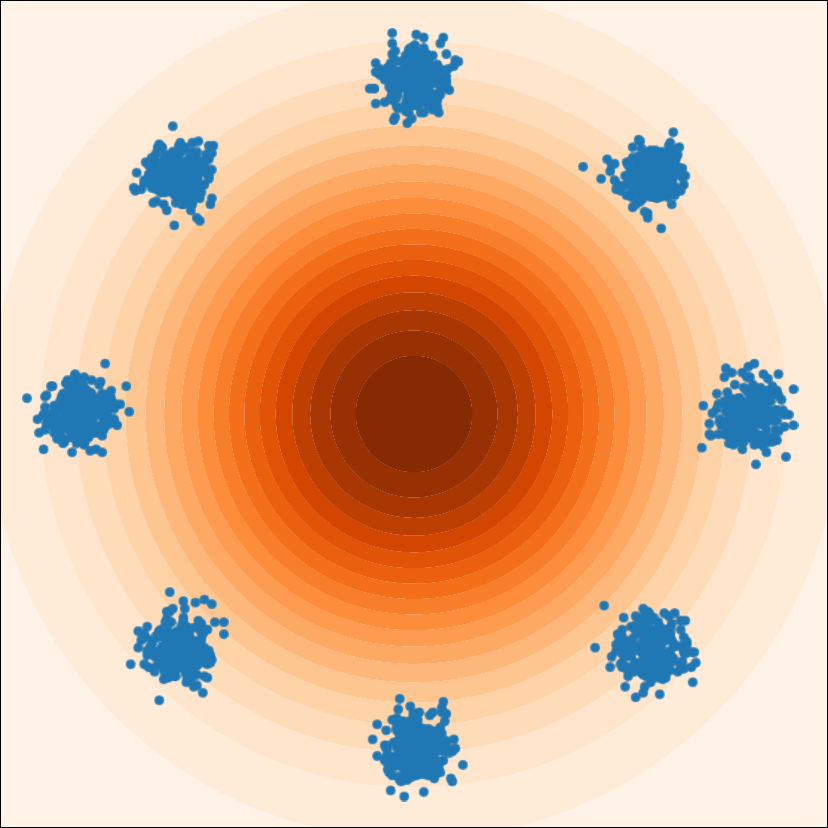}}
\subfigure{\includegraphics[trim = 0mm 0mm 0mm 0mm,  clip,width=0.48\linewidth ]{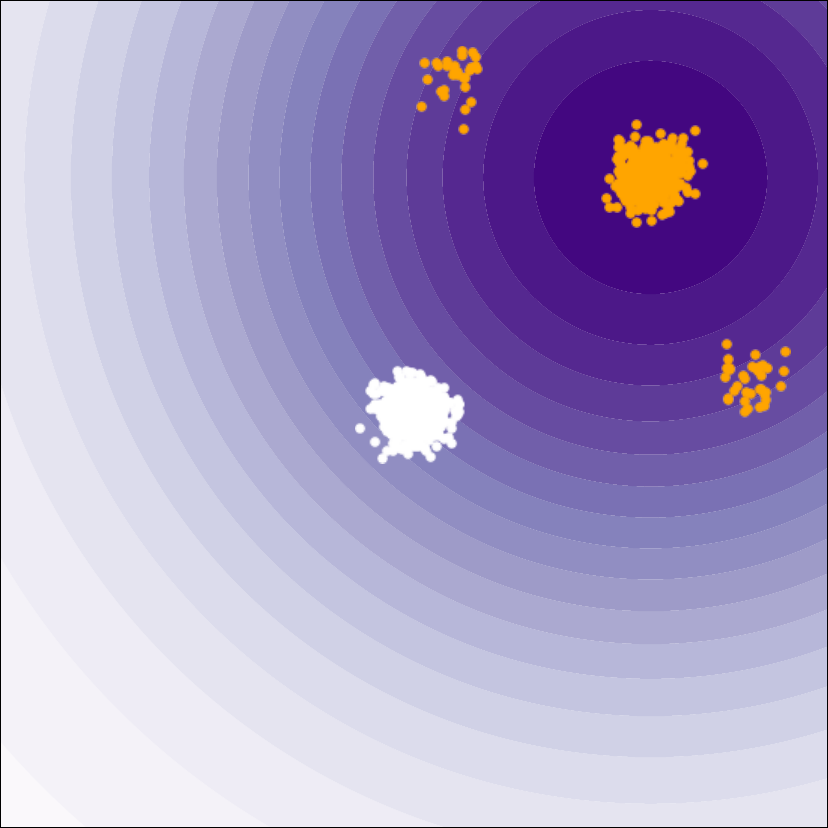}}
\caption{\textbf{Illustration of SWFP performance on a bandit toy example: } (left) source distribution and behavior data in the replay buffer; (right) implicit SWFP steers the source distribution toward the high-performance behavior policy, heatmap shows target policy. }
\label{fig:toys}
\end{wrapfigure}  

To evaluate the ability of our discretised step flow model to recover a target distribution, we begin with a synthetic two-dimensional example.
The experiment starts with a source distribution as a standard Gaussian centered at the origin, and the behavior data distribution as an eight-component Gaussian mixture (from which 1,000 samples are drawn).  The right of Fig.~\ref{fig:toys} visualizes the distribution evolution produced by a six-step SWFP ($N=6$) as it transports the source distribution towards the target (indicated by the purple level sets).
White particles initially sampled from the Gaussian source are gradually transported out of the low-reward annulus and into the high-reward modes, closely following the gradient field of the target policy.

\subsection{Efficient Online Adaptation}
\begin{figure*}[t]
	\begin{center}
		\subfigure{\includegraphics[trim = 4mm 0mm 0mm 0mm,  clip,width=0.32\linewidth ]{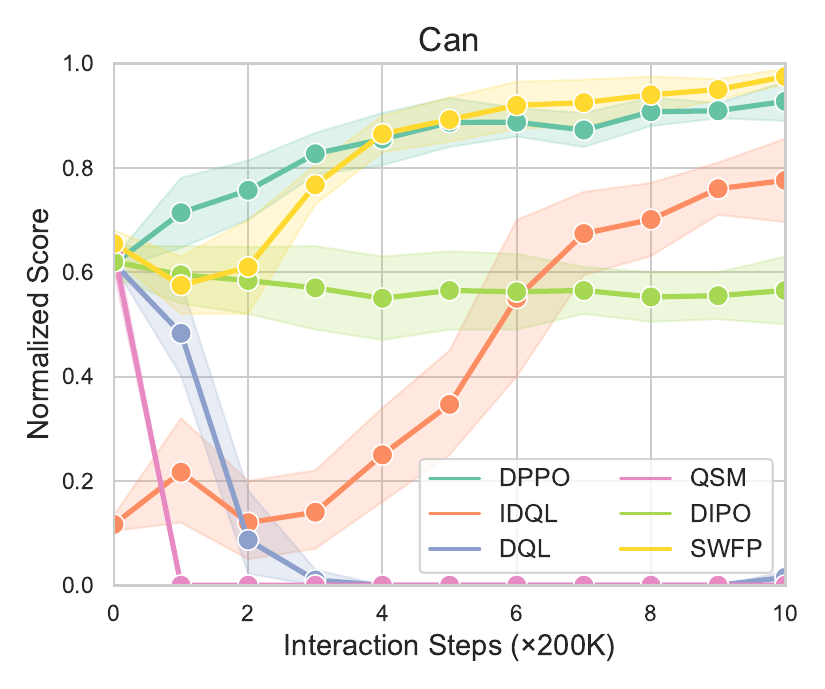}}
		\subfigure{\includegraphics[trim = 4mm 0mm 0mm 0mm,  clip,width=0.32\linewidth ]{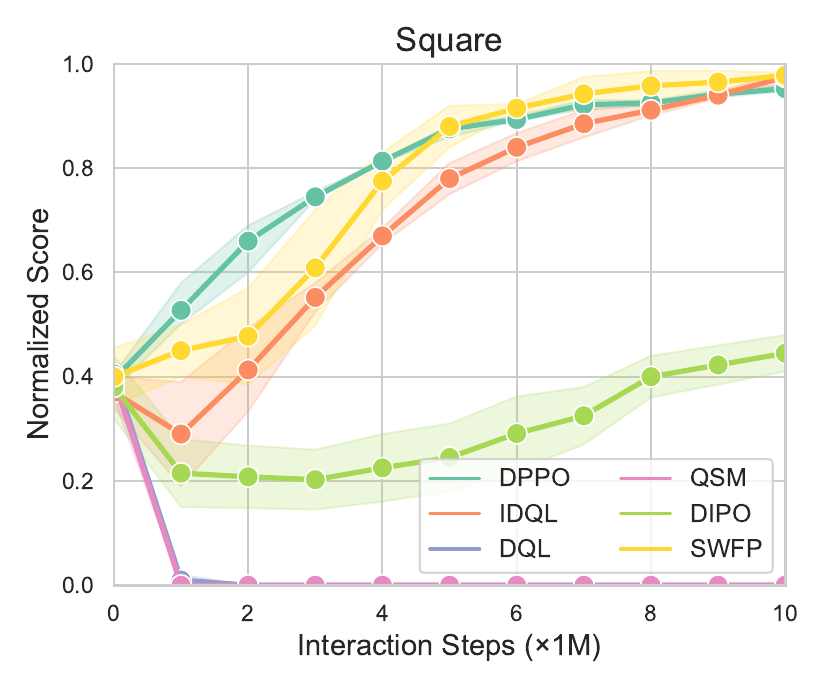}}
		\subfigure{\includegraphics[trim = 4mm 0mm 0mm 0mm,  clip,width=0.32\linewidth ]{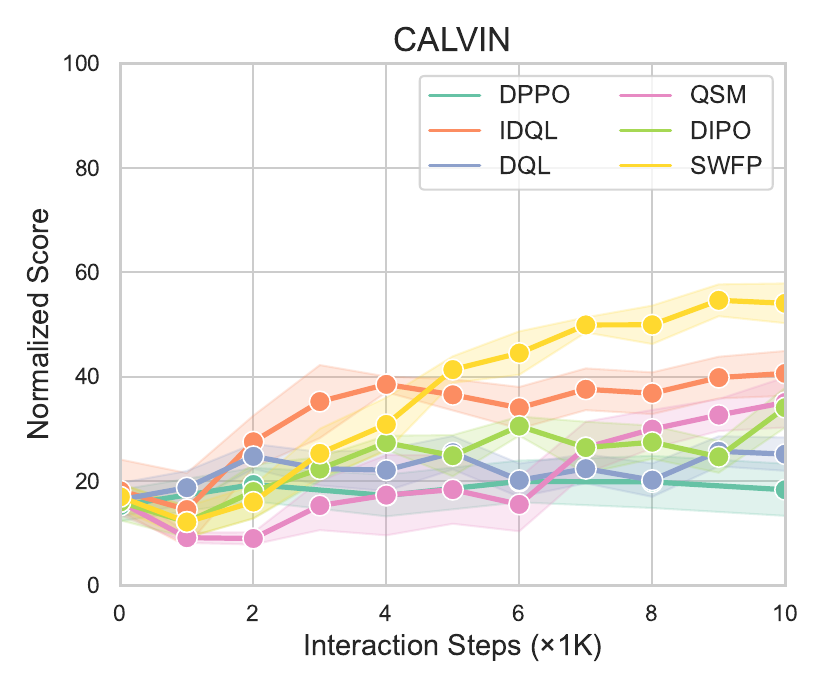}}
		\subfigure{\includegraphics[trim = 4mm 0mm 0mm 0mm,  clip,width=0.32\linewidth ]{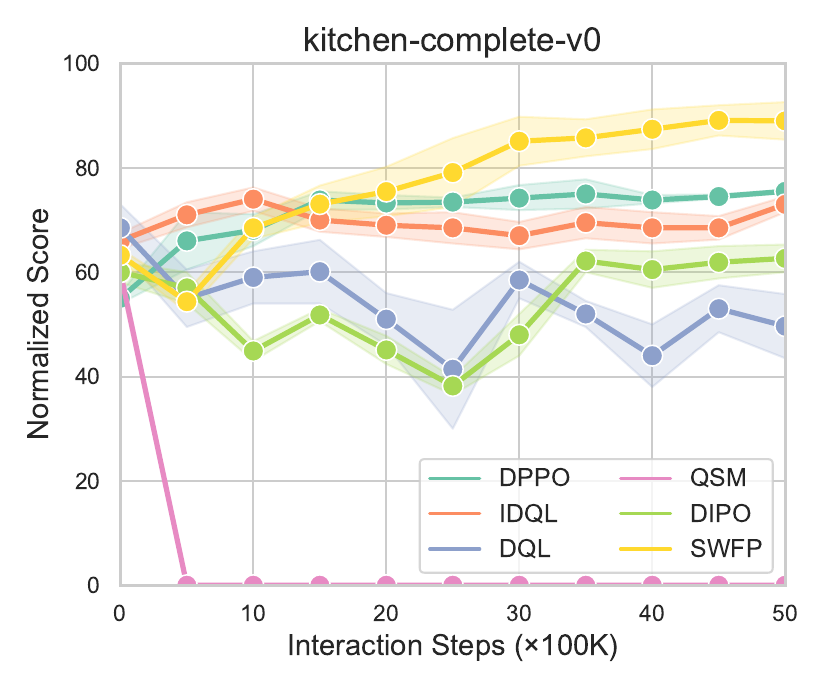}}
		\subfigure{\includegraphics[trim = 4mm 0mm 0mm 0mm,  clip,width=0.32\linewidth ]{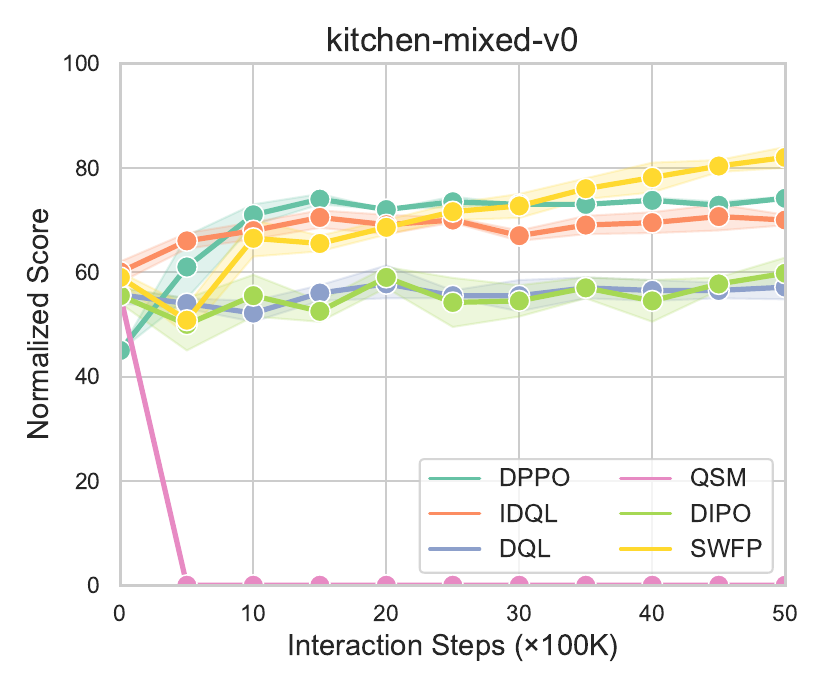}}
		\subfigure{\includegraphics[trim = 4mm 0mm 0mm 0mm,  clip,width=0.32\linewidth ]{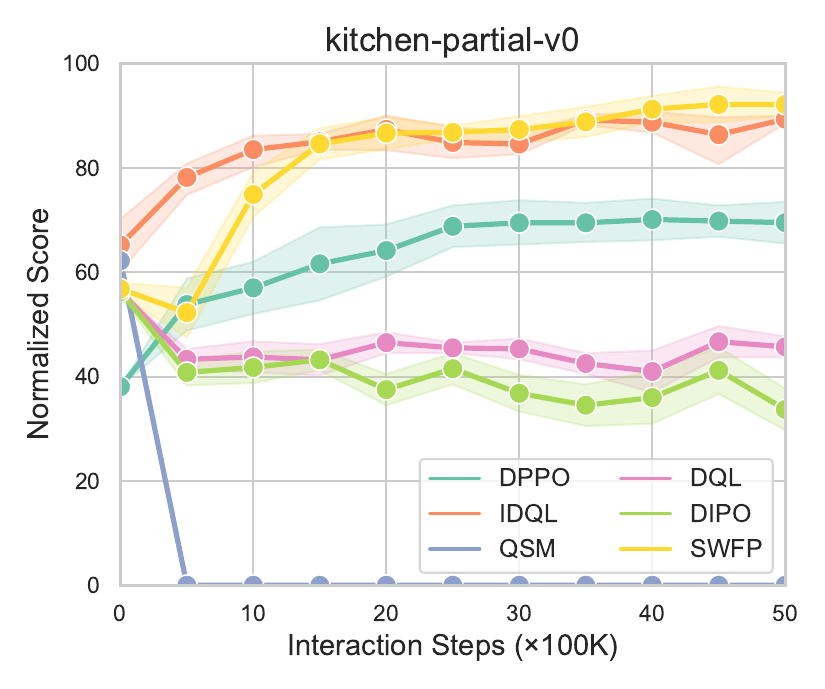}}
		\caption{\textbf{Learning curves of online fine-tuning with various methods.} Observe that SWFP largely always dominates or attains similar performance to the next best method. }
		\label{fig:comparisons}
	\end{center}
    \vskip -0.3in
\end{figure*}
We first examine the online adaptation setting, where a diffusion or flow policy pretrained on demonstrations is adapted using online samples.
Overall, SWFP consistently and significantly improves the fine-tuning efficiency and asymptotic performance of flow policies.
 Notably, SWFP maintains high normalized scores across diverse tasks like kitchen-complete-v0, CALVIN, and Can, a stability that contrasts sharply with the considerable fluctuations observed in baselines like DQL and IDQL. 
This superior robustness is directly traceable to SWFP's core design. By utilizing the Stepwise JKO Iteration, SWFP ensures that each policy update is intrinsically regularized by the Wasserstein metric. This Wasserstein Trust Region strictly limits the ``magnitude of change" in the policy distribution, effectively mitigating catastrophic divergence and compounding errors often seen in off-policy methods operating in high-dimensional continuous action spaces. Furthermore, SWFP breaks the optimization into a chain of short, easily trained flow blocks. This avoids the high computational and stability costs of backpropagating through a monolithic, multi-step flow. In essence, SWFP unifies the expressive power of flow policies with the principled stability of JKO proximal optimization, resulting in a robust and highly effective online fine-tuning approach.

\begin{table*}[t]
  \centering
  \caption{Comparison of SWFP with other offline-to-online RL algorithms. SWFP outperforms every other approach, both in terms of the offline performance (left of $\rightarrow$) and performance after fine-tuning (right of $\rightarrow$). }
  \vskip 0.1in
    \begin{tabular}{cccccc}
    \toprule
    \toprule
    \multirow{2}[2]{*}{} & \multicolumn{3}{c}{Franka-Kitchen} & \multicolumn{2}{c}{RoboMimic} \\
          & Complete-v0 & Mixed-v0 & Partial-v0 & Can-State   & Square-State \\
    \midrule
    RLPD  &   $0 \rightarrow 18$    &   $0 \rightarrow 14$    &  $0 \rightarrow 34$     &  $0 \rightarrow 0$     & $0 \rightarrow 3$ \\
    \quad Cal-QL \quad \ & $19 \rightarrow 57$ & $37 \rightarrow 72$ & $59 \rightarrow 84$ & $ 0 \rightarrow 0 $      & $0 \rightarrow 0$ \\
 %    IQL   & $54 \rightarrow 60$ & $48 
 % \rightarrow 48$ & $57 \rightarrow 50$ & $ 34 \rightarrow 30$      & $28 \rightarrow 15 $ \\
    IBRL  &  $0 \rightarrow 25$     &   $0 \rightarrow 13$    &  $0 \rightarrow 15$    & $0 \rightarrow 64$    & $0 \rightarrow 50$ \\
    \midrule
    % OTPR-U &  $61 \rightarrow $     & $59 \rightarrow 79$      &   $42 \rightarrow 93$    & $63 \rightarrow 99$   &  $40 \rightarrow 98$\\
    SWFP  &  $  63 \rightarrow 89$     & $ 59 \rightarrow 82$      &   $ 57 \rightarrow 92$    & $ 65 \rightarrow 97$   &  $40 \rightarrow 98$\\
    \bottomrule
    \bottomrule
    \end{tabular}%
  \label{tab:exp1RL}%
  \vskip -0.1in
\end{table*}%
\subsection{Efficient Offline-to-Online Adaptation}
We investigate the effectiveness of SWFP in the challenging offline-to-online setting, comparing its performance against established methods like RLPD~\cite{ball2023efficient}, Cal-QL~\cite{nakamoto2024cal}, and IBRL~\cite{hu2023imitation}, all of which leverage offline data for off-policy updates. We evaluate these methods on Franka-Kitchen and RoboMimic environents. All of results are shown on Table~\ref{tab:exp1RL}. 
Across the Franka-Kitchen domains, SWFP consistently demonstrates superior adaptation capability, achieving impressive score improvements, notably rising from 63 to 89 in Kitchen-Complete-v0 and leading decisively with 92 in Kitchen-Partial-v0—scores where baselines like RLPD and IQL perform significantly worse. This success highlights SWFP's effective mechanism for refining a pretrained flow policy via online interaction. Furthermore, in the RoboMimic environment, SWFP continues to excel, achieving high scores of 97 in Can-State and 98 in Square-State, showcasing its robustness across diverse, high-fidelity scenarios. Although IBRL performs the best among the non-flow competitors, a significant performance gap remains, solidifying SWFP as the state-of-the-art approach for fine-tuning expressive flow policies in the offline-to-online paradigm.

\begin{wrapfigure}{r}{0.45\textwidth}
\centering
\includegraphics[trim = 0mm 0mm 0mm 0mm,  clip,width=0.98\linewidth ]{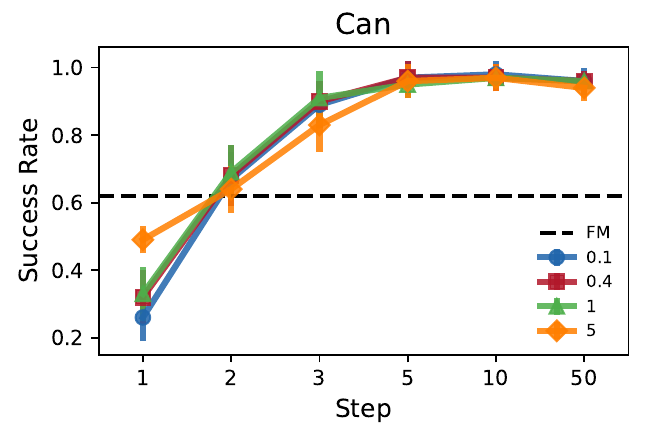}
\caption{Sensitivity of Hyperparameters.}
\label{fig:ablation}
\end{wrapfigure}  
\subsection{Sensitivity Analysis}
This section investigates the sensitivity of our algorithm to the discrete step number $N$ and the Wasserstein-2 scale. Evaluation on the RoboMimic-Can task over 10 runs yields the results in Figure~\ref{fig:ablation}. The figure reveals two main insights: First, increasing the step number leads to a consistent improvement in success rate until performance plateaus at a high level, with a significant jump observed after $N=5$. Second, and more importantly, the four curves representing different Wasserstein-2 scales $(0.1, 0.4, 1, 5)$ cluster closely together. This indicates that the W-2 trust-region is not a sensitive parameter, as the algorithm maintains strong performance across a wide range of values, consistently outperforming the pre-trained FM baseline (dashed line).

\section{Conclusion}

In this work, we introduce the Stepwise Flow Policy (SWFP) framework to enable stable and efficient online fine-tuning of flow-based policies. By aligning the discretized flow-matching process with the JKO scheme from optimal transport, SWFP decomposes policy improvement into a sequence of regularized, local updates in probability space. This approach ensures stable adaptation through Wasserstein trust regions and entropic regularization. Extensive experiments demonstrate that SWFP achieves superior stability and performance compared to existing methods, providing a principled and practical path for online reinforcement learning with generative policies.

\bibliography{iclr2025_conference}
\bibliographystyle{iclr2025_conference}

\newpage
\appendix

\section{Proof of Proposition~\ref{p1}}

\begin{proof}
The first claim of Proposition~\ref{p1} states that the iterated policy $\pi(\mathbf{a} |\mathbf{s})$ converges to $p_{s,\pi}(\boldsymbol{a}) \propto e^{Q(\boldsymbol{a}, \boldsymbol{s})}$. Recall that the iterative flow defined in\ref{eq:jkopi} with energy functional in \ref{eq:fe} is equivalent to a Fokker-Planck equation:
\[
\partial_n \pi_n = \nabla \cdot (-\pi_n\nabla Q(\boldsymbol{a}, \boldsymbol{s}))  + \nabla^2 \pi_n,
\]
where $n$ denotes the iteration index. The Fokker–Planck operator on
the right-hand side admits a unique invariant measure.
Setting $\partial_{n}\pi_n=0$ yields
\[
  \nabla\!\cdot\!\Bigl(
        -\,\pi_\infty\,\nabla Q
      \Bigr)
  \;+\;
  \nabla^{2}\pi_\infty
  \;=\;0.    
\]

A standard calculation shows that the unique normalisable solution is the
Gibbs measure
\[
   \pi_\infty(\boldsymbol{a}\,|\,\boldsymbol{s})
   \;=\;
   \frac{1}{Z(\boldsymbol{s})}
   \exp\!\bigl(Q(\boldsymbol{a},\boldsymbol{s})\bigr),
\qquad
   Z(\boldsymbol{s})=\int
     \exp\!\bigl(Q(\boldsymbol{a},\boldsymbol{s})\bigr)\,
     \mathrm d\boldsymbol{a}.
\]
Hence, as $n\to\infty$, the sequence $\{\pi_n\}$ converges to
$p_{s,\pi}$.

% The unique invariant
% probability measure for the FP equation is:
% \begin{equation}
%     \lim_{n\rightarrow\infty}\pi_n = e^{Q(\boldsymbol{a}, \boldsymbol{s})}
% \end{equation}

% Now we derive the soft Bellman equation. At time step $t$, following from the definition of $Q(\boldsymbol{s}, \boldsymbol{a})$,
% \begin{align*}
% Q(\boldsymbol{s}_t, \boldsymbol{a}_t) &= r(\boldsymbol{s}_t, \boldsymbol{a}_t) + \mathbb{E}_{(\boldsymbol{s}_{t+1}, \boldsymbol{a}_{t+1}, \cdots) \sim (\rho_\pi, \pi)} \sum_{l=1}^{\infty} \gamma^l r(\boldsymbol{s}_{t+l}, \boldsymbol{a}_{t+l}) \\
% &= r(\boldsymbol{s}_t, \boldsymbol{a}_t) + \gamma \mathbb{E}_{\boldsymbol{s}_{t+1} \sim \rho_\pi} \mathbb{E}_{\boldsymbol{a}_{t+1} \sim \pi} \Bigg[ r(\boldsymbol{s}_{t+1}, \boldsymbol{a}_{t+1}) \\
% &\quad + \mathbb{E}_{(\boldsymbol{s}_{t+2}, \boldsymbol{a}_{t+2}, \cdots) \sim (\rho_\pi, \pi)} \sum_{l=1}^{\infty} \gamma^l r(\boldsymbol{s}_{t+1+l}, \boldsymbol{a}_{t+1+l}) \Bigg]
% \end{align*}

% Since $\pi(\boldsymbol{s} \mid \boldsymbol{a}) = e^{Q(\boldsymbol{s} \mid \boldsymbol{a}) - V_\pi(\boldsymbol{s})}$ where $V_\pi(\boldsymbol{s}) = \int_{\mathcal{A}} Q(\boldsymbol{s}_{t+1}, \boldsymbol{a}) \mathrm{d}\boldsymbol{a}$, we have
% \[
% \begin{aligned}
% Q(\boldsymbol{s}_t, \boldsymbol{a}_t) &= r(\boldsymbol{s}_t, \boldsymbol{a}_t) \\
% &\quad + \gamma \mathbb{E}_{\boldsymbol{s}_{t+1} \sim \rho_\pi} \left[ V_\pi(\boldsymbol{s}_{t+1}) - \mathcal{H}(\pi(\cdot \mid \boldsymbol{s}_{t+1})) \right]
% \end{aligned}
% \]

For any policy $\pi$ the soft state–action value is defined as
\[
   Q_\pi(\boldsymbol{s}_t,\boldsymbol{a}_t)
   \;=\;
   r(\boldsymbol{s}_t,\boldsymbol{a}_t)
   +\;
   \mathbb{E}_{(\boldsymbol{s}_{t+1:t+\infty},\,
                \boldsymbol{a}_{t+1:t+\infty})\sim(\rho_\pi,\pi)}
   \Bigl[
        \sum_{l=1}^{\infty}\gamma^l
        r(\boldsymbol{s}_{t+l},\boldsymbol{a}_{t+l})
   \Bigr].
\]
Isolating the first transition yields
\begin{align*}
   Q_\pi(\boldsymbol{s}_t,\boldsymbol{a}_t)
   &= r(\boldsymbol{s}_t,\boldsymbol{a}_t)  \\
   &\quad
   +\;
   \gamma\,\mathbb{E}_{\boldsymbol{s}_{t+1}\sim\rho_\pi}\,
        \mathbb{E}_{\boldsymbol{a}_{t+1}\sim\pi}\!
        \Bigl[
             r(\boldsymbol{s}_{t+1},\boldsymbol{a}_{t+1})
             +\!
             \mathbb{E}_{(\boldsymbol{s}_{t+2:\infty},\,
                          \boldsymbol{a}_{t+2:\infty})}
             \sum_{l=1}^{\infty}\gamma^l
             r(\boldsymbol{s}_{t+1+l},\boldsymbol{a}_{t+1+l})
        \Bigr].
\end{align*}

Introduce the (soft) state value
$
   V_\pi(\boldsymbol{s})
   =\mathbb{E}_{\boldsymbol{a}\sim\pi}
     \bigl[\,Q_\pi(\boldsymbol{s},\boldsymbol{a})\bigr],
$
and recall that the Boltzmann policy can be written as  
\(
   \pi(\boldsymbol{a}\,|\,\boldsymbol{s})
   = \exp\bigl(Q_\pi(\boldsymbol{s},\boldsymbol{a})-V_\pi(\boldsymbol{s})\bigr).
\)
Using this form and the definition of (differential) Shannon entropy  
$
   \mathcal{H}\!\left(\pi(\cdot\mid\boldsymbol{s})\right)
   = -\mathbb{E}_{\boldsymbol{a}\sim\pi}
       \bigl[\log\pi(\boldsymbol{a}\mid\boldsymbol{s})\bigr],
$
we can rewrite the inner expectation:
\[
   \mathbb{E}_{\boldsymbol{a}_{t+1}\sim\pi}
     \bigl[Q_\pi(\boldsymbol{s}_{t+1},\boldsymbol{a}_{t+1})\bigr]
   = V_\pi(\boldsymbol{s}_{t+1})
     - \mathcal{H}\!\left(\pi(\cdot\mid\boldsymbol{s}_{t+1})\right).
\]

Substituting this result into the previous expansion gives the soft Bellman equation
\[
   Q_\pi(\boldsymbol{s}_t,\boldsymbol{a}_t)
   =
   r(\boldsymbol{s}_t,\boldsymbol{a}_t)
   +\gamma\,
     \mathbb{E}_{\boldsymbol{s}_{t+1}\sim\rho_\pi}
     \Bigl[
        V_\pi(\boldsymbol{s}_{t+1})
        -\mathcal{H}\!\left(\pi(\cdot\mid\boldsymbol{s}_{t+1})\right)
     \Bigr].
\]
\begin{flushright}
$\blacksquare$
\end{flushright}
\end{proof}

\section{More Analysis}\label{app:ana}
The per-step training objective, defined by Equation~\ref{eq:jkopi}, can be viewed as the addition of two key components: the variational objective (measuring the closeness of the pushforwarded density to the target policy) and the Wasserstein regularization term. This \( \mathcal{W}_2 \) term plays a critical role by serving as a {proximal operator} that actively regularizes the ``amount of movement'' enforced by the transport map \( F_{n,\theta} \). Intuitively, among all transport maps that can successfully reduce the \(\mathrm{KL}\) divergence to the target policy, the regularization selects the one that achieves the result with the smallest possible movement. By penalizing excessive displacement between consecutive policy iterates, the \( \mathcal{W}_2 \) term encourages straighter, more stable transport paths in the probability space~\citep{xie2025flow}. This stabilization inherently reduces the risk of mode collapse and may decrease the total number of neural network blocks required to reach the final target distribution. Critically, in a particle-based implementation of the \( n \)-th flow block, this \( \mathcal{W}_2 \) term becomes computationally tractable, directly corresponding to the average squared movement over \( m \) particles along the \(\mathrm{ODE}\) trajectory: \( \frac{1}{m} \sum_{i=1}^m \| x^i(t_n) - x^i(t_{n-1}) \|^2 \).

The convergence guarantee for the $\text{SWFP}$ is established by mirroring the analysis of W2-proximal gradient descent ($\mathrm{GD}$), leveraging the assumption that each $\text{JKO}$ minimization step is solved with an approximation error of $O(\varepsilon)$. By utilizing the $\lambda$-convexity of the $\mathrm{KL}$-divergence functional, $\mathcal{F}(\rho) := \mathrm{KL}(\rho\|q)$, in the Wasserstein space, we derive the Evolution Variational Inequality~\citep{muratori2020gradient,xie2025flow}:
$$\left(1 + \frac{\tau\lambda}{2}\right) \mathcal{W}^2(p_{n+1}, q) + 2\tau \left(\mathcal{F}(p_{n+1}) - \mathcal{F}(q)\right) \le \mathcal{W}^2(p_{n}, q) + O(\varepsilon^2), $$
which is a key step that confirms the exponential convergence of the Wasserstein $\mathrm{GD}$, ensuring the policy iterates $p_n$ quickly approach the target distribution $q$. Specifically, this analysis shows that a $\mathrm{KL}$-divergence error of $O(\varepsilon^2)$ is achieved after executing only approximately $\log(1/\varepsilon)$ $\text{JKO}$ steps, demonstrating the theoretical efficiency of $\text{SWFP}$ in recovering the target policy.

\section{Details for Experiments}
All experiments are conducted on an NVIDIA Tesla A100 80GB GPU, and all fine-tuning methods use the same pre-trained policy. In the pretraining, the observations and actions are normalized to $[0, 1]$ using min/max statistics from the pre-training dataset. No history observation (pixel, proprioception, or ground-truth object states) is used. The deffusion policy and flow policy is trained with learning rate $1e-4$ decayed to $1e-5$ with a cosine schedule, weight decay 1e-6 and 50 parallelized. For Franka-Kitchen and Robomimic tasks, epochs is $8000$ and batch size is $128$; for CALVIN tasks, epochs is $5000$ and batch size is $512$.

\subsection{Details and Hyper-parameters for SWFP}

SWFP has two most important hyperparameters: the discretized stepsize $\tau$ and the temperature $\alpha$. Also note that we can only evaluate the gradient of W2 up to a constant, there needs to a parameter balancing the gradient of the energy functional $F$ and the Wasserstein term. We denote this hyparameter as $\epsilon$. In the experiments, if not explicitly stated, the default setting for these parameters are $\epsilon = 0.4$. For temperature parameter, we performed a grid search over $\alpha \in \{2,3,4,5,6,7,8\}$.

% Table generated by Excel2LaTeX from sheet 'Sheet1'
\begin{table}[htbp]
  \centering
  \caption{Hyper-parameters for SWFP}
  \setlength{\tabcolsep}{2pt}
    \begin{tabular}{ccccc}
    \toprule
    \toprule
    \multirow{2}[2]{*}{Parameter} & \multicolumn{4}{c}{Task} \\
          & Franka-Ketichen & CALVIN & Robomimic-Can & Robomimic-Square \\
    \midrule
    Buffer Size  & 1000000 & 250000 & 250000 & 250000 \\
    Actor Learning Rate & 1.00E-05 & 1.00E-05 & 1.00E-05 & 1.00E-05 \\
    Discount Factor & 0.99 & 0.99 & 0.999 & 0.999 \\
    Optimizer & \multicolumn{4}{c}{Adam} \\
    ODESolver & \multicolumn{4}{c}{Midpoint Euler} \\
    Block Number  $N$ & 5 & 5 & 5 & 5 \\
    $\alpha$ & 5 & 5 & 4 & 4 \\
    $\tau$ & 0.1 & 0.1 & 0.1 & 0.1 \\
    Actor Batch Size & 1024  & 1024  & 1024  & 1024 \\
    Hidden Layer Sizes  & [512, 512, 512] & [512, 512, 512] & [256, 256, 256] & [256, 256, 256] \\
    Q Batch Size & 256   & 256   & 256   & 256 \\
    \bottomrule
    \bottomrule
    \end{tabular}%
  \label{tab:addlabel}%
\end{table}%

\subsection{Details and Hyper-parameters for Baselines}

\textbf{DPPO} For the state-based tasks Robomimic and FrankaKitchen, we trained DPPO-MLP following the original paper's specifications, using an action chunking size of 4 for Robomimic and 8 for FrankaKitchen. For the pixel-based task CALVIN, we trained DPPO-ViT-MLP with an action chunking size of 4.

\textbf{IDQL} We employ the IDQL-Imp version of IDQL, wherein the Q-function, value function, and diffusion policy are refined through new experiences. For Robomimic tasks, we employ the same network architecture as DPPO, while the original IDQL architectures are preserved for Franka-Kitchen and CALVIN. For the IQL $\tau$ expectile, we set it to 0.7 for each task. 

\textbf{DQL} We set the weighting coefficient to $0.5$ for Robomimic, $0.005$ for Franka-Kitchen and $0.01$ for CALVIN.

\textbf{IBRL} We adhere to the original implementations' hyperparameters, with wider (1024) MLP layers and dropout during pre-training.

\textbf{Cal-QL} We set the mixing ratio to $0.25$ for Franka-Kitchen and $0.5$ for CALVIN and Robomimic.

\textbf{QSM} We set the weighting coefficient $\alpha$ to 10.

\textbf{DIPO} We set the weighting coefficient $\alpha$ to $1e-4$ and hyperparameter $M$ to $10$.

\begin{algorithm}[htb]
\renewcommand{\algorithmicrequire}{\textbf{Require:}}
\renewcommand{\algorithmicensure}{\textbf{Output:}}
   \caption{Online Policy Training (SWFP)}
   \label{alg:traingpolicybrief}
\begin{algorithmic}[1]
   \REQUIRE The pre-trained flow policy $\pi^\theta$, replay buffer $\mathcal{B} = \varnothing$, initial Q-function $Q^\varphi$, MDP $\mathcal{M}$.
   % \STATE {\bfseries Output: } Trained conditional score-based policy $s_\theta$.
   \STATE Target parameters:  $\bar{\theta} \leftarrow \theta$,  $\bar{\varphi} \leftarrow \varphi$
   \FOR{each epoch $k$ in  $\{1, 2,  \dots, K\}$}
   \STATE \textcolor{blue}{\% Sampling and experience replay.}
   \STATE Interact with $\mathcal{M}$ using the policy $\pi^{\bar\theta}$.
   \STATE Update replay buffer $\mathcal{B}$.
   \STATE Sample minibatch $(\boldsymbol{s},\{\boldsymbol{a}_{n}\}^N_{n = 0},{r},\boldsymbol{s}') $ from $ \mathcal{B}$.
   \STATE \textcolor{blue}{\% Update Q function}
   \STATE Compute empirical values $\hat V^{\bar\varphi}(\boldsymbol{s}') $.
   \STATE Update $\varphi$ with the empirical gradient $\hat\nabla_\varphi J_Q(\varphi) $.
   \STATE \textcolor{blue}{\% Update policy}
   \FOR{each policy improvement step}
        \STATE Sample $n$ uniformly from $\{1, 2, \dots, N\}$.
        \STATE Compute $\mathcal{W}^2_2(\pi^{\theta}_n,\pi^{\bar\theta}_{n-1})$ and $\mathcal{W}^2_2(\pi^{\theta}_{n}, \pi^{\bar\theta}_n)$.
        \STATE Compute the empirical gradient $\hat\nabla_\theta J_{\pi_n}(\theta)$
        \STATE Update prior policy parameters $\bar{\theta}$.
   \ENDFOR
   \STATE Update target Q-function parameters $\bar\varphi$.
   \ENDFOR
\end{algorithmic}
\end{algorithm}

\end{document}